# The Optimization of the Constant Flow Parallel Micropump Using RBF Neural Network

Chenyang Ma
University of Michigan
Department of EECS
Michigan, U.S.
e-mail: dannymcy@umich.edu

Boyuan Xu, Hesheng Liu
Shanghai Wufeng Scientific Instrument Co., Ltd.
R&D Department
Shanghai, China
e-mail: {jamesxu, hsliu}@wufengtech.com

*Abstract*—The objective of this work is to optimize the performance of a constant flow parallel mechanical displacement micropump, which has parallel pump chambers and incorporates passive check valves. The critical task is to minimize the pressure pulse caused by regurgitation, which negatively impacts the constant flow rate, during the reciprocating motion when the left and right pumps interchange their role of aspiration and transfusion. Previous works attempt to solve this issue via the mechanical design of passive check valves. In this work, the novel concept of overlap time is proposed, and the issue is solved from the aspect of control theory by implementing a RBF neural network trained by both unsupervised and supervised learning. The experimental results indicate that the pressure pulse is optimized in the range of 0.15 - 0.25 MPa, which is a significant improvement compared to the maximum pump working pressure of 40 MPa.

## I. Introduction

Micropumps are the type of pump to manipulate small amount of fluid, and usually with precisions calibrated to nanoliter, microliter, or milliliter. As the essential component of microfluidics systems, micropumps are widely used in fields such as biological sampling, medical systems, and cooling of microelectronics [1-4]. Micropumps have different designs and implementations, and there are a few novel designs in recent years such as peristaltic polydimethylsiloxane (PDMS) micropump, four phase AC electroosmotic micropump, and MEMS-based valveless piezoelectric micropumps [5-7]. Traditional micropumps can be broadly classified into two main categories. They are mechanical displacement micropumps, which exert pressure on the fluid, and dynamic micropumps, which use energy to provide the fluid either momentum or pressure [4, 8, 9]. Reciprocating micropumps are the most common displacement pumps, and they are based either on diaphragm or plunger with the incorporation of different types of valves and pump chamber designs [8]. Among all the reciprocating pumps, in terms of chamber design, serial micropumps and parallel micropumps with passive check valves are the two mainstreams. Compared to serial micropumps, parallel pumps have the advantages of higher efficiency and extended machine life because their working principle depends on reciprocating motion in which the left and right pumps conduct motion alternatively, and interchanges their role of aspiration and transfusion [8].

A constant flow micropump is a type of micropump that maintains constant flow rate of the working fluid during operation, and it has crucial applications such as medical diagnostics and as an essential part of precise analysis instruments [10]. For constant flow micropumps with parallel pump chambers, passive check valves are normally incorporated, and the pump designs are widely studied. Parallel micropump which adopts check valves with a tethered-plate structure [11] is designed and implemented with specific application for integrated chemical analyzing systems. Planar micropump [12, 13] with parallel pump chambers is researched in which diffuser/nozzle is used to replace check valves. Micropump manufactured by combined thermoplastic molding and membrane techniques with parallel pump chambers [14] is researched. Several sensors are combined with micropumps and valves for the chemical analysis of fluids and gases.

For constant flow parallel micropump, one critical issue is to maintain the constant flow rate when performing operations on the fluids [8, 15, 16], and check valves play an important role. The check valves permit the flow in one direction when the pumps connected in parallel interchange their role of aspiration and transfusion. During the process, the issue of regurgitation is commonly presented inside the valve. Pump working pressure is defined as the pressure exerted by the pump chamber on the fluid surface area during transfusion. The regurgitation causes a drop in working pressure, or a pressure pulse, when the pump conducts transfusion, which disturbs the constant flow rate of the transmitting fluid.

Numerous researches are conducted on the optimization of constant flow parallel micropump, and most of them focus on the design of passive check valves, attempting to improve the stability, reliability, ease of manufacture of the check valves, and increase the maximum working pressure. Check valves with mechanical flap structure [17-19] are researched to stabilize the pressure pulse. Invariants of the flap check valves are also researched such as check valves based on cantilever structures [20] and in-plane flap check valves [21]. Check valves with a tethered-plate structure [11, 22, 23] are researched to maximize the flow rate and working pressure. Ball check valves [24, 25] are researched in an attempt to find the correlation between the performance of the micropump and the working frequency.

Previous works solve the issue of regurgitation and pressure pulse via the mechanical design of the passive check valves. In this paper, rather than focusing on the mechanical design, the concept of overlap time is proposed to stabilize the pressure pulse. By modeling the pump system and identifying important parameters, RBF neural network trained with a combination of unsupervised learning of center locations and scaling factors by K-means clustering and supervised learning of weights and bias or center locations using Adam [36] as the optimizer is implemented to find the desired overlap time. RBF neural network is widely used to model complex nonlinear systems because of its high generalization power and strong tolerance to input noise [37]. RBF neural network based on an improved artificial bee colony (ABC) algorithm is used on traffic flow prediction. RBF neural network which adopts minimum parameter learning method to replace weights and bias to improve the convergence speed is implemented to model the electromagnet levitation control system of maglev trains [27]. A parallel radial basis function neural network in conjunction with chaos theory is implemented on the failure detection of hydraulic pump [28].

The structure of this paper can be classified as follows. First, the designed micropump system using distributed control system is reviewed, and the mechanical structure of individual parallel micropumps is examined. Then, the issue of regurgitation in the passive check valve and how it impacts the working pressure is explained. After that, the parallel micropump is modeled and important parameters are identified. The reasons to fit the system using RBF neural network is then explained from the aspect of the uniqueness of training data points, and a combination of unsupervised and supervised training is used to train the network. Finally, the experimental setting is explained and the final performance is evaluated.

We, to the best of our knowledge, is the first to optimize and maintain the constant flow rate of a parallel mechanical displacement micropump from the aspect of control theory. The main contributions of our study can be concluded as the following:

1. We propose the novel concept of overlap time to stabilize the pressure pulse.
2. Rather than focusing on the mechanical design of the passive check valves, we solve the issue from the aspect of control theory using RBF neural network, which saves time and cost of production.
3. We propose and train RBF neural network with a combination of unsupervised and supervised training methods.
4. After experimental testing, the pressure pulse is optimized in the range of 0.15 - 0.25 MPa, which is a significant improvement compared to the benchmark of 10 - 30 MPa.

## II. MICROPUMP SYSTEM

### A. System Overview and Mechanical Structure

The designed micropump system consists of the injection pump modules, Modbus router, and the host computer as illustrated in Fig. 1. Each injection pump module is made of a four-level injection pump module controlled by the Modbus router. Equivalently, individual injection pump has four types of configurations as shown in Table I.

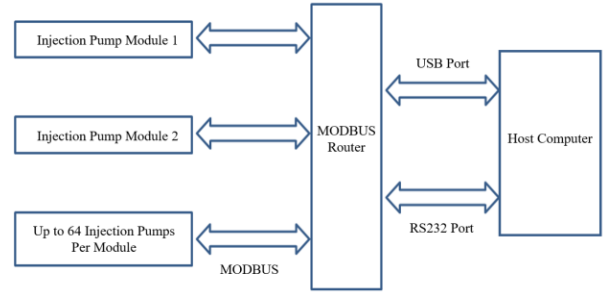

Figure 1. Overview of micropump system modules

TABLE I. CONFIGURATIONS OF MICRO INJECTION PUMP

| Pump Type | Volume | Maximum Working Pressure | Maximum Flow Rate |
|---|---|---|---|
| 1 | 25μL | 40 MPa | $3.0 \times 10^3$ μL/min |
| 2 | 125μL | 40 MPa | 7.6 mL/min |
| 3 | 500 μL | 40 MPa | 15.0 mL/min |
| 4 | 5 mL | 10 MPa | 600.4 mL/min |

Fig. 2 illustrates the appearance and structure of four types of micro injection pumps. Fig. 3 illustrates the structure of the injection pump module. Each micro injection pump is driven by linear stepping motors and has two position sensors. One position sensor is set at zero position while the other one limits the position. Multiple switch valves and passive check valves are assembled in the module. Pressure sensor, temperature sensor, and leakage sensor are incorporated into the pump. The pressure sensor monitors the working pressure and classifies it into either normal, low, or high. Typically, abnormal low pressure is caused by the malfunction and shut down of the injection pumps, and abnormal high pressure is caused by clog in the pipe and nozzle. The temperature sensor monitors the working status of motors and the temperature of the transmitting fluids. The leakage sensor detects any leak in the pipes.

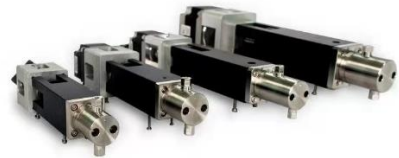

Figure 2. Appearance and structure of four types of micro injection pump

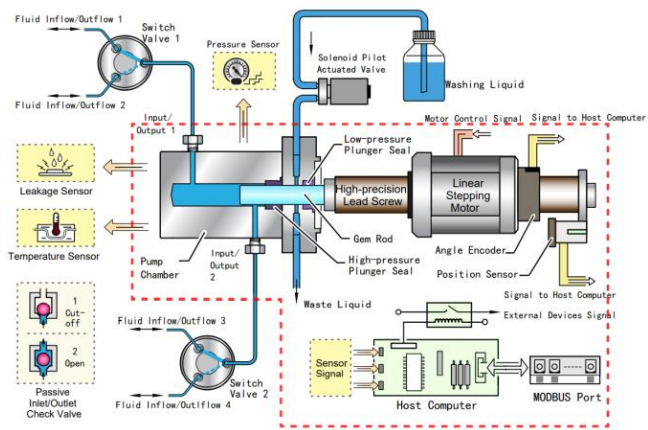

Figure 3. Structure of injection pump module

Each injection pump module can control up to 64 injection pumps. Based on the application requirements, each module's configuration is manipulated in the host computer.

### B. Distributed Control System

The micropump system uses distributed control system, and its controller consists of the host computer and ARM embedded control panel. As illustrated in Fig. 4, using GUI, the host computer manipulates working time, working pressure, and working sequence of each injection pump module, and configures time, temperature, flow amount, and flow rate requirements. By manipulating injection time and selecting inputs and outputs of the injection pump modules, the micro pump control system is capable to perform operations on multiple solvents at both constant and changing flow rates while satisfying the requirements of set flow amount, set time, and required precision.

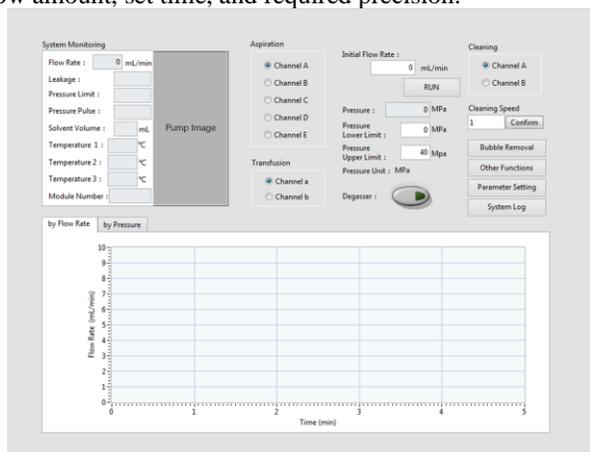

Figure 4. GUI of Micropump Control System

In general, the distributed control system with four-level micro injection pump module and multiple sensors realizes the automatic switch between different modules, and is capable to inject multiple solvents simultaneously with precisions calibrated to nanoliter, microliter, and milliliter.

## III. MAINTAINING PARALLEL PUMP CONSTANT FLOW RATE

### A. The Structure of Parallel Pumps

Within the injection pump module, injection pumps with the same configuration can be assembled as both parallel and series pump to perform fluid injection at constant flow rate. Fig. 5 illustrates the structure of the assembled parallel pump.

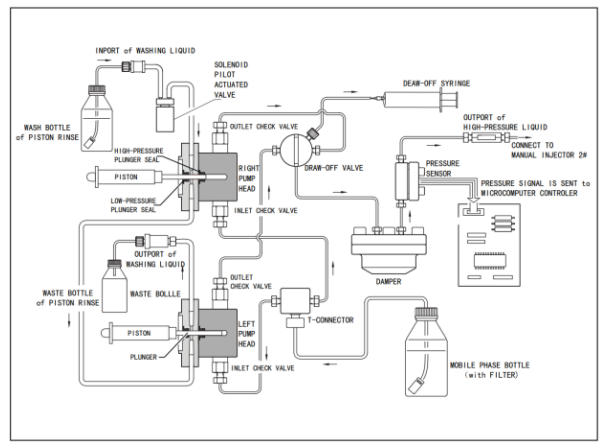

Figure 5. Structure of the parallel micropump

The reciprocating double plunger parallel pump adopts a compact inner structure and its integrated structure adopts modularized design, which consists of two pumps symmetrical at both left and right, pump driven modules including stepping motor and motor-driven plate, membrane style damper, pressure sensor, leakage sensor, post-column cleaning unit, etc. Fluid flows through inlet check valve from solvent bottle, entering into liquid feed pump, which consists of pump head, pump seat, inlet and outlet check valve, plunger seal, and plunger rod. The stepping motor driven by the host computer controls the left and right pumps in antiphase, conducting reciprocating motion to alternatively aspirate and transfuse the fluid. The pair of pumps consistently interchange their role during the movement, and the point at which they interchange their role is called the shifting point. The fluid flowing into the left and right pumps meet at draw-off valve, and the pressure pulse created during the aspiration and transfusion in the left and right pumps is partially removed by the damper. The pressure is then measured by the pressure sensor and sent to the host computer.

### B. The Structure of Check Valve and Its Impaction on Working Pressure and Flow Rate

Fig. 6 illustrates the essential components of the structure of both inlet and outlet check valve. The check valve is a specially designed ball check valve, and the gem ball stays at either the gland ring or gem ball seat inside the valve depending on the flow direction, and seals the valve seat when there is no flow or when reverse flow occurs.

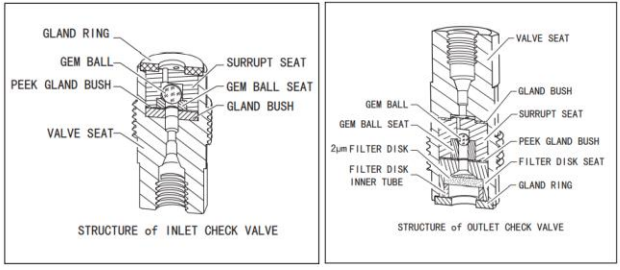

Figure 6. Structure of the ball check valve

During transfusion, flow is permitted, and high-pressure transmitting fluid flows in the direction of gland bush to gland ring. Usually, a pressure around 20 MPa is exerted by the fluid, which indicates there is a force of 20 N on an area of 1 mm$^2$ inside the check valve. Since the pressure around the gem ball seat is higher than the pressure around the gland ring, the gem ball stays at the gland ring.

At the shifting point when the conducting motion changes from transfusion to aspiration, the transmitting fluid, which enters from the gland bush, is cut off. The gem ball drops to the gem ball seat and seals the flow. During this process, there is a sudden and large change in both the direction and magnitude of the fluid pressure inside the check valve. The pressure around the gland ring and gland bush is imbalanced, so the gem ball bounces a few times, and needs a few milliseconds to return to complete stationery and stays at the gem ball seat. During bouncing, a certain amount of regurgitation is induced, and its amount is positively correlated to the pump working pressure.

During transfusion, high working pressure is exerted by the pump chamber on the transmitting fluid. The regurgitation results in a sudden drop on the working pressure of the pump starting from the shifting point, and lasts for a short amount of time, as shown in Fig. 7.

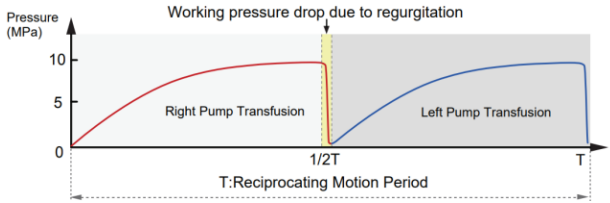

Figure 7. Drop in working pressure during transfusion

As a result, this pressure pulse causes a pulse on the flow rate of the transmitting fluid with shape similar to Fig. 7. The constant flow rate of the transmitting fluid is disrupted.

IV. METHODOLOGY

A. Overlap Time

Fig. 8 illustrates the theoretical motion of the left and right pumps during the interchange of aspiration and transfusion. The motion changes instantaneously at the shifting point, and pressure pulse will be induced in the real implementation. The solution proposed is to create a period of overlap time.

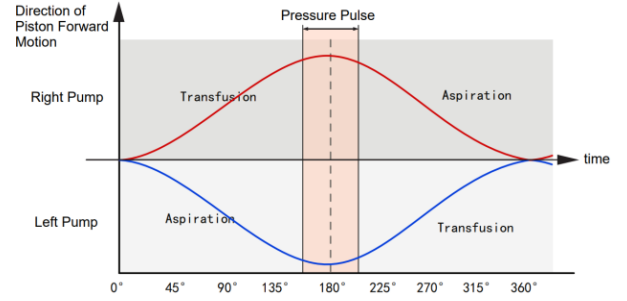

Figure 8. Theoretical motion of the left and right pumps

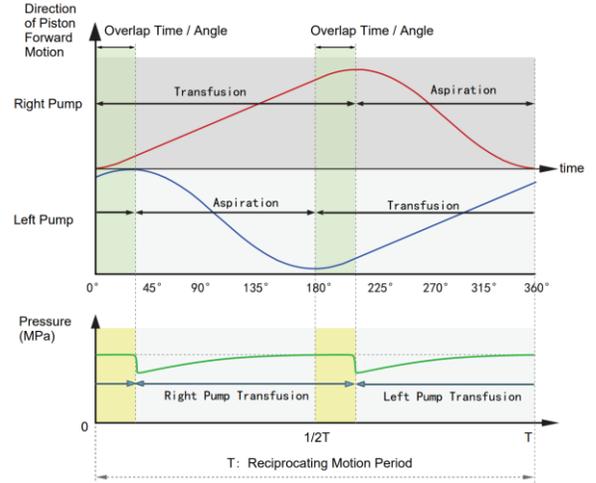

Figure 9. Overlap time to compensate the shifting point

As illustrated in the upper half of Fig. 9, in one cycle of motion, instead of having the same amount of time to aspirate and transfuse fluid in the left and right pumps, both pumps lessen the time to aspirate fluid and extend the time to transfuse fluid. In a small amount of time, the left and the right pumps collectively conduct transfusion. As the motion conducted by the right pump changes from transfusion to aspiration, the forward velocity of the right pump gradually decreases. On the contrary, as the motion conducted by the left pump changes from aspiration to transfusion, the forward velocity of the left pump gradually increases. Consequently, as illustrated in the lower half of Fig. 9, the working pressure of the right pump is just slightly higher than the working pressure of the left pump at the shifting point when they interchange their conducting motion. Compared to the working pressure drop illustrated in Fig. 7, the use of overlap time to extend the transfusion time greatly compensates the regurgitation and pressure pulse. As a result, the sudden pulse on the flow rate of the transmitting fluid is dampened.

To determine the overlap time, the starting point and the duration need to be known. This can be done by thinking from the perspective of overlap angle. For the starting point, it always takes place at 180°. For the duration, it is simple to calculate it from the overlap angle by using:

$$t_{op} = T \frac{\theta_{op}}{360} \qquad (1)$$

where, $t_{op}$ is the overlap time, $\theta_{op}$ is the overlap angle in degrees, and T is the period.

The issue presented is to determine the optimal overlap angle as it is extremely hard to model the entire system. The regurgitation is affected by various factors, and the system is nonlinear. Thus, RBF neural network is proposed to find the overlap angle.

### B. System Modelling

The overlap angle is affected by multiple observable and measurable variables.

1. The dynamic viscosity of the transmitting fluid at room temperature (20 °C), denoted as $\mu$, is known as the injected fluid is determined before the experiment.

2. The working pressure when the left and right pumps interchange their role of aspiration and transfusion, denoted as $P$. This can be known by reading the pressure signal sent to the host computer by the pressure sensor installed above the damper.

3. Rotational speed of the plunger, denoted as $\omega$, is closely related to the flow rate set by the users in the GUI. The rotational speed of the plunger can be expressed as:

$$\omega = \frac{Q_{max}}{A_{plu}D_m} \quad (2)$$

where, $\omega$ is the rotational speed of the plunger, $Q_{max}$ is the set flow rate of the transmitting fluid by the host computer, $A_{plu}$ is the cross-sectional area of the plunger, and $D_m$ is the distance moved by the motor per revolution. In (2), $A_{plu}D_m$ calculates the flow amount per revolution.

4. The flow rate of the fluid when passing through the check valve, which should be distinguished from the set flow rate of the transmitting fluid by the host computer. It can be expressed as:

$$Q = 2\lambda_z \lambda_f \frac{A_p V_m S}{T_a + T_t} \quad (3)$$

where, $Q$ is the flow rate passing through the check valve, $\lambda_z$ is the correction coefficient of pressure, $\lambda_f$ is the correction coefficient of flow rate, $A_p$ is the cross-sectional area of the pump, $V_m$ is the stepping motor's displacement per step, $S$ is the total number of motor steps in one cycle of reciprocating motion, $T_a$ is the period of aspiration, and $T_t$ is the period of transfusion. $A_p V_m S$ calculates the volume in the pump, and $T_a + T_t$ is the period, which is equivalent to T in (1).

In (3), $\lambda_z$ and $\lambda_f$ can be further expressed as:

$$\lambda_z = 1 + \frac{Z}{100}\frac{P}{P_{max}} \quad (4)$$

$$\lambda_f = 1 \pm \frac{F}{100} \quad (5)$$

where, $Z$ is the correction constant of pressure, $F$ is the correction constant of flow rate, $P$ is the pump working pressure, and $P_{max}$ is the maximum pump working pressure.

$Z$ is related to the amount of regurgitation under high working pressure, and $F$ is related to the amount of regurgitation caused by imprecisions in the mechanical parts such as imprecision of pump volume and imprecision of stepping motor's displacement per step. Both of the correction constants have the unit of mL min$^{-1}$, and their exact values are calibrated in the experiments, which will be explained in Section V. B.

The above identified variables serve as inputs to the RBF neural network.

### C. Radial Basis Function Neural Network

Radial basis function (RBF) neural network is a three-level forward network that is capable to fit nonlinear systems. It is chosen to model the micropump system due to its strong tolerance to input noises and good generalization for relatively small dataset [37], which is true for the dataset collected in this case, and will be explained in Section V. B. and Section V. C. in details. RBF neural network consists of the input layer, hidden layer, and output layer, and generalized structure of the network is shown in Fig. 10:

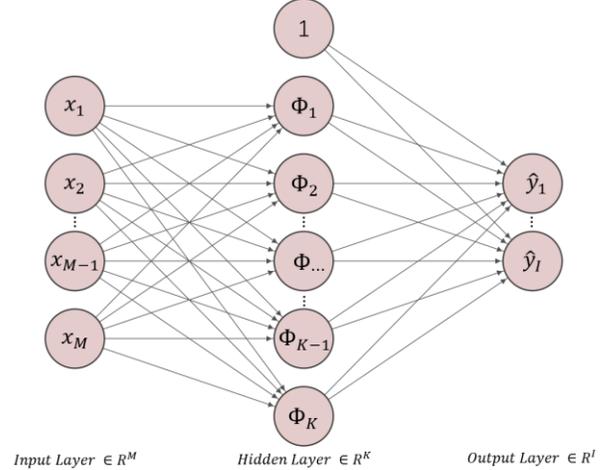

Figure 10. Generalized structure of the RBF neural network

The input layer is a vector of an arbitrary number of variables, and each node in the hidden layer contains a parameter vector called center or prototype. The output of the input layer is calculated by evaluating the dissimilarity between the input vector and the center of each node in the hidden layer. Then, this value is fed to the activation function, which is the radial basis function. Finally, the output of the hidden layer is sent to the output layer, and the final output is calculated using weights and bias. The output layer is equivalent to a fully connected layer. The RBF neural network can be generally expressed as:

$$\hat{y}_i = \sum_{j=1}^{K} w_{ji} \Phi(\|x - c_j\|) + \beta_i \quad (6)$$

where,
$x$ is the input vector of arbitrary size $M$,
$K$ is the number of nodes in the hidden layer, $j \in \{1,\dots,K\}$,

$I$ is the number of nodes in the output layer, $i \in \{1,\ldots,I\}$,
$\Phi$ is the radial basis function,
$w_{ji}$ is the weight of the $j^{th}$ hidden node and $i^{th}$ output,
$\beta_i$ is the bias of the $i^{th}$ output node,
$C_j$ is the $j^{th}$ center in the hidden layer,
$\hat{y}_i$ is the output of the $i^{th}$ output node.

As a common practice, the dissimilarity between the input vector and the centers in the hidden layer is calculated by the Euclidian norm [29-31], and Gaussian function is chosen to be the radial basis function $\Phi$, which can be expressed as:

$$\varphi(\|x - c_j\|) = \exp\left(-\frac{\|x - c_j\|^2}{2\sigma^2}\right) \tag{7}$$

where $\sigma$ is the scaling factor or the width.

### D. Combination of Unsupervised and Supervised Training of the RBF Neural Network

For a RBF neural network, the determination of the number of nodes in the hidden layer, the clustering method of the centers, the adjustment of the scaling parameter of the Gaussian function, and the training of the weights and bias in the output layer are all crucial [29-31], and will affect the outcomes of training. There are a few important aspects to look into.

First, the number of nodes in the hidden layer is closely related to the generalizing capability and the complexity of the RBF neural network [29, 30]. If the number of nodes is insufficient in the hidden layer, the learning ability of the network will be lessened. On the contrary, if the number of nodes is in excess, there will be issue of poor generalization and overfitting. The number of nodes can either be fixed before the training, or adjusting during the training when locating the centers by using backpropagation.

In addition, the clustering method, which determines the value of the centers, also affects the final output [29, 31, 32]. The clustering method can be divided into two categories [31, 32]. The first category only considers the distribution of training inputs, while the second category considers both inputs and outputs when clustering.

For the first category, common clustering methods include:
1. Choosing a set of grid points, which requires a large number of basis functions for inputs data with high dimensional space.
2. Randomly select the centers.
3. Unsupervised clustering algorithms such as K-means clustering, decision trees, learning vector quantization (LVQ), and their invariants.

For the second category, methods include:
1. Unsupervised clustering algorithm based on both inputs and outputs.
2. Supervised learning of the centers by backpropagation.

3. On-line learning in which the centers are learned by backpropagation, and the number centers can be finetuned.

The scaling parameter of the Gaussian function, $\sigma$, also affects the transition of the RBF network [29, 31]. A too small $\sigma$ will result in the overfitting of the data set. Conversely, a too large $\sigma$ will result in an overly smoothed probability density.

Finally, the training and optimization of the weights and bias between the hidden layer and the output layer affect the final performance of the RBF network. Popular approaches include using numerical method such as least squares, or backpropagation with optimizers [29, 30, 32, 33].

In the experiment context, a combination of unsupervised training and supervised training is adapted for the RBF neural network. The general pipeline can be concluded in Fig. 11:

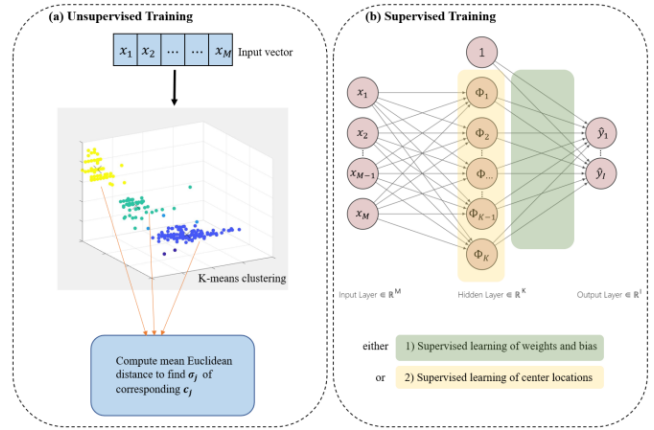

Figure 11. General pipeline of training of the RBF neural network

In the unsupervised training, K-means clustering that takes both inputs and outputs into consideration is adapted to locate the centers. The empirical quantization error can be expressed as:

$$E(c_1, \ldots, c_K) = \sum_{j=1}^{K} \sum_{x^u \in r_j} \|x^u - c_j\| \tag{8}$$

where $x^u$ is the $u^{th}$ input vector, $c_1,\ldots,c_k$ is the clustering centers with the corresponding clusters $r_1,\ldots,r_k$. The goal is to minimize the quantization error of each cluster. The algorithm can be concluded in the following steps:

1. K initial centers are randomly chosen.

2. Each input is assigned to one of the K centers with the lowest dissimilarity calculated by the Euclidean distance.

$$x^u \in r_j(n), \text{if } j(x^u) = \min_j \|x^u - c_j(n)\|, j = 1, \ldots, K \tag{9}$$

where $c_j(n)$ denotes the $j^{th}$ center in the $n^{th}$ iteration, and $r_j(i)$ denotes the corresponding $j^{th}$ cluster in the $n^{th}$ iteration.

3. K clustering centers are recalculated according to (10):

$$c_j(n+1) = \frac{1}{|r_j(n)|} \sum_{x \in r_j(n)} x \qquad (10)$$

where $|r_j(n)|$ is the total number of data in the cluster $r_j(n)$.

4. If in two consecutive iterations, all the data in each cluster $r_j$ for $j=1,\ldots,K$ are equal. Equivalently,

$$\sum_{j=1}^{K} r_j(n+1) = r_j(n) \qquad (11)$$

$$\sum_{j=1}^{K} c_j(n+1) = c_j(n) \qquad (12)$$

If the above two equations hold, the algorithm is finished and the centers are successfully found. On the contrary, go to step 2 and start the next iteration.

After performing the K-means clustering, the scaling factors, $\sigma$, are calculated for each center $c_j$. For each $c_j$, the corresponding $\sigma_j$ can be expressed as:

$$\sigma_j = \frac{1}{|r_j|} \sum_{x^u \in r_j} \|x^u - c_j\| \qquad (13)$$

which is the mean Euclidean distance of all the data in the cluster $r_j$ to the center $c_j$.

In the part of supervised learning, either weights and bias or center locations are trained using backpropagation instead of both. The most vital reason to use supervised learning to train the weights and bias is the input variables are not equally important. Both the supervised learning of the weights and bias and the learning of center locations can make some features in the inputs data more important than others, achieving better performance. However, only the learning of one of them is supervised because adjusting both weights and bias and center locations will cause redundancy, which is more likely to create local minima [31]. The supervised training of the scaling factor is not adapted because it is proven to be not effective in most cases [31, 33-35]. In Section V., the performance of the two approaches is compared.

After supervised training, the output of the RBF network can be expressed as:

$$\hat{y} = [\varphi(\|x^u - c_1\|) \quad \varphi(\|x^u - c_2\|) \quad \cdots \quad \varphi(\|x^u - c_k\|) \quad 1] \cdot \begin{bmatrix} w11 \\ w21 \\ \vdots \\ wk1 \\ \beta 1 \end{bmatrix}$$

(14)

Or equivalently,

$$\hat{Y} = V \cdot W \qquad (15)$$

$W$ has one column because the only output, the overlap angle, is a scaler. Backpropagation method using Adam as the optimizer is implemented, and its goal is to minimize the error function:

$$E = \sum (\hat{Y} - Y)^2 \qquad (16)$$

where $Y$ is the ground truth.

For supervised learning of the weights and bias, the center locations and scaling factors are the results of the K-means clustering. In the simplest form, the update of the weights and bias in each backpropagation is:

$$w_{ji} = w_{ji} - \eta \frac{\partial E}{\partial w_{ji}} \qquad (17)$$

where $\eta$ is the step size.

For the supervised learning of the center locations, the error function is the same as (16). The initialization center locations are the results of the K-means clustering, and the scaling factor is updated in each learning epoch according to (13). The update of the centers in each backpropagation is:

$$c_j = c_j - \eta \frac{\partial E}{\partial c_j} \qquad (18)$$

and each time the centers are updated, the weights and bias are calculated by the conventional least squares method, which can be expressed as:

$$\widehat{W} = (V^T V)^{-1} V^T \hat{Y} \qquad (19)$$

where $W$, $V$, and $\hat{Y}$ are the same as (15).

## V. EXPERIMENT

### A. Experimental Setting

A normal flow rate of the transmitting fluid is in the range of 0.1 - 5 mL min$^{-1}$. Maximum flow rate can achieve 10 mL min$^{-1}$, and will be used in rare cases. Normal testing fluids include water ($H_2O$), methanol ($CH_3OH$), and acetonitrile ($CH_3CN$), with respective dynamic viscosity, $\mu$, of 1.002 cP, 0.594 cP, and 0.389 cP at 20°C. The value of experimental parameters is listed in Table II.

TABLE II. VALUE OF EXPERIMENTAL PARAMETERS

| | |
|---|---|
| The volume of the pump ($A_p V_m S$) | 125 μL |
| Cross-sectional area of the plunger ($A_{plu}$) | 7.917 mm$^2$ |
| Stepping motor's displacement per step ($V_m$) | 0.01 mm |
| Stepping motor's displacement per revolution ($D_m$) | 2 mm |
| Total number of motor steps in one cycle of reciprocating motion ($S$) | 1580 |
| Maximum Pump Working Pressure ($P_{max}$) | 40 MPa |
| Maximum speed of stepping motor | 2000 step s$^{-1}$ |

### B. Determining the Value of Correction Constant

A damper tube with very small radius and high internal resistance is used to simulate the pressure drop caused by regurgitation by exerting back pressure on the working fluid. A damper tube is assembled next to the pressure sensor as the replacement of check values, as shown in Fig. 12.

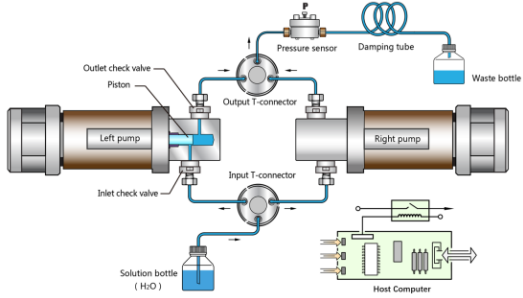

Figure 12. Incorporation of damper tube in the pump system

By adjusting the length of the damper tube, the value of back pressure can be configured in the range of 0 - 40 MPa.

$F$, the correction constant of flow rate, is related to inaccuracies in the mechanical parts, and is calibrated under low back pressure. $Z$, the correction constant of pressure, is related to high pump working pressure, and is calibrated under high back pressure. The values of other parameters are fixed. The fluid is chosen to be water. A standard flow rate of 1 mL min$^{-1}$ is configured in the host computer, and four cycles of reciprocating motion are needed to initiate the pump system to reach the flow rate of 1 mL min$^{-1}$. The period of the reciprocating motion, T, is set at 10 s.

A low pressure drop is considered to be in the range of 0 - 5 MPa. The value of $F$, the correction constant of flow rate, is determined first. The back pressure is fixed at 2 MPa. The pump system operates for 10 minutes, and the volume of fluid in the waste bottle is measured. The value of $F$ can be obtained by averaging the measured volume over time.

The value of $Z$, the correction constant of pressure, is then determined. The length of the damper tube is adjusted, and the back pressure is controlled in the range of 5 - 40 MPa. The pump system operates for 10 minutes, and the value of $Z$ can be obtained by averaging the measured volume over time.

Table III concludes the value of $F$ and $Z$ obtained at different back pressure.

TABLE III. ACQUIRED VALUE OF $F$ AND $Z$

| Back Pressure (MPa) | $F$ (mL min$^{-1}$) | $Z$ (mL min$^{-1}$) |
|---|---|---|
| 0 - 5 | 13 | 0 |
| 5 - 10 |  | 6 |
| 10 - 15 |  | 7 |
| 15 - 20 |  | 15 |
| 20 - 25 |  | 18 |
| 25 - 30 |  | 25 |
| 30 - 35 |  | 31 |
| 35 - 40 |  | 42 |

Under 5 MPa, $F$, the correction constant of flow rate, is dominating, so $Z$, the correction constant of pressure, is trivial and taken to be 0 mL min$^{-1}$ in (4). With pressure greater than 5 MPa, both $Z$ and $F$ are taken into consideration in (5).

### C. Acquiring the Training Data Points

The experimental pump system is the same as the one presented in Fig. 12. To acquire the training data points, the flow rate is changed in the range of 0.1 - 5 mL min$^{-1}$, and the back pressure in the damper tube is controlled in the range of 1 - 40 MPa. Testing fluids include water ($H_2O$), methanol ($CH_3OH$), and acetonitrile ($CH_3CN$). The period of reciprocating motion of the pump is altered in the range of 3 - 15 s. Using (2) - (5), important variables identified in Section IV. B. including dynamic viscosity, working pressure, rotational speed of the plunger, and flow rate in check valve can be calculated.

After a combination of the four variables is determined, the overlap angle is gradually adjusted and tested, starting from 5 degrees to 45 degrees. The pump system operates for 10 minutes, and the volume of fluid in the waste bottle is measured and recorded. The optimum overlap angle is taken to be the one with the minimum averaged fluid volume.

A total of 500 data points is acquired and partitioned into training set, validation set, and test set of 400, 50, and 50 data points.

### D. Evaluation Metric

Instantaneous pressure drop at the shifting point is measured by pressure sensor, which is sent to the host computer is used as the measuring matrix. Flow rate is not used as the evaluation metric because as explained in Section II. A. and Section III. A., there is no flowmeter assembled in the micropump system.

### E. Experimental Results

Experiments are conducted on the test set. Benchmarks include using 1) no overlap angle and 2) a fixed overlap angle of 30 degrees. Proposed methods include using overlap angle from RBF neural network trained by 1) supervised learning of weights and bias and 2) supervised learning of center locations. Table IV concludes the center locations acquired by unsupervised learning of K-means clustering and supervised learning using the results of K-means clustering as initialization. Each center is represented in the form of [dynamic viscosity (cP), working pressure (MPa), rotational speed of the plunger (min$^{-1}$), flow rate in check valve (mL min$^{-1}$), tested overlap angle (°)].

TABLE IV. CENTER LOCATIONS

| Center Number | Centers by K-means Clustering | Centers by Supervised Learning |
|---|---|---|
| 1 | [0.6678, 26.9527, 163.8395, 3.0249, 32.0356] | [0.6752, 23.6358, 149.8346, 2.3074, 26.1745] |
| 2 | [0.6777, 20.7523, 60.5342, 2.6517, 29.4673] | [0.6446, 17.8564, 81.8963, 2.5668, 26.7785] |
| 3 | [0.4300, 14.4655, 277.2065, 2.4001, 25.0829] | [0.6788, 20.6220, 250.5065, 2.9405, 29.5985] |
| 4 | [0.6680, 21.5704, 216.2636, 2.5292, 32.3736] | [0.7468, 19.6113, 196.2653, 3.3383, 23.9041] |
| 5 | [0.8318, 35.0733, 295.1115, 2.8560, 40.1815] | [0.6230, 31.8682, 280.9736, 2.1264, 37.9274] |

For each center location, the acquired overlap angle is averaged. Table V concludes the acquired and averaged overlap angle from trained RBF neural networks using two different supervised training methods:

TABLE V. OVERLAP ANGLE FROM RBF NEURAL NETWORKS

| Center Number | Overlap Angle by Supervised Learning of Weights and Bias (°) | Overlap Angle by Supervised Learning of Center locations (°) |
|---|---|---|
| 1 | 32.0356 | 29.5985 |
| 2 | 29.4673 | 26.7785 |
| 3 | 25.0829 | 26.1745 |
| 4 | 32.3736 | 23.9041 |
| 5 | 40.1815 | 37.9274 |

For comparison of two benchmarks and two proposed methods, data points are clustered by the result of K-means clustering, and the instantaneous pressure drop at the shifting point is measured by conducting experiments on the micropump using no overlap angle, a fixed overlap angle of 30 degrees, and the acquired overlap angle from RBF neural networks using two training methods. The measured pressure pulse is then averaged at each cluster. Table VI concludes the result with illustration in Fig. 13:

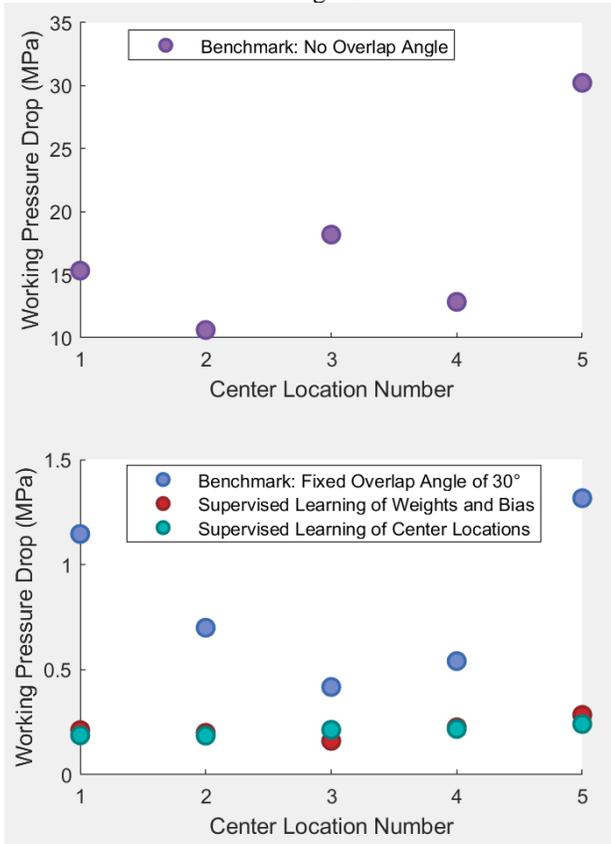

Figure 13. Comparison of two benchmarks and two proposed methods on working pressure drop

TABLE VI: COMPARISON OF TWO BENCHMARKS AND TWO PROPOSED METHODS ON WORKING PRESSURE DROP

| Center Number | Averaged Pressure Drop with No Overlap Angle (MPa) | Averaged Pressure Drop with Fixed Overlap Angle of 30 Degrees (MPa) | Averaged Pressure Drop with Supervised Learning of Weights and Bias (MPa) | Averaged Pressure Drop with Supervised Learning of Center Locations (MPa) |
|---|---|---|---|---|
| 1 | 15.3264 | 1.146 | 0.2122 | 0.1873 |
| 2 | 10.6159 | 0.6995 | 0.1995 | 0.1857 |
| 3 | 18.1815 | 0.4175 | 0.1609 | 0.2137 |
| 4 | 12.8454 | 0.5408 | 0.2257 | 0.2177 |
| 5 | 30.2286 | 1.3164 | 0.2847 | 0.2412 |

The benchmark of using a fixed overlap angle of 30 degrees has significant improvement over the benchmark of using no overlap angle by lowering the pressure pulse in the range of 10 - 30 MPa to 0.5 - 1.5 MPa. Using the proposed training methods of RBF neural network, both supervised learning of weights and bias and supervised learning of center locations greatly outperform the benchmark of using a fixed overlap angle of 30 degrees. The supervised learning of center locations performs generally better than the supervised learning of weights and bias, and optimizes the pressure pulse in the range of 0.15 - 0.25 MPa.

## VI. CONCLUSION

In this work, we optimize the issue of pressure pulse that happened during transfusion of the reciprocating motion of a constant flow parallel mechanical displacement micropump from the aspect of control theory. By using RBF neural networks trained with a combination of unsupervised and supervised methods, experimental results show that the pressure pulse is optimized in the range of 0.15 - 0.25 MPa, which is a significant improvement compared to the maximum pump working pressure of 40 MPa.

ACKNOWLEDGMENT

This work was supported by Shanghai Wufeng Scientific Instrument Co., Ltd.